CAPÍTULO IV

# Grandes modelos de lenguaje: ¿de la predicción de palabras a la comprensión?*

**Carlos Gómez-Rodríguez**

Los grandes modelos de lenguaje, como el conocido ChatGPT, han supuesto una inesperada revolución en el ámbito de la inteligencia artificial. Por un lado, cuentan con multitud de aplicaciones prácticas y un enorme potencial todavía por explorar. Por otro lado, son también objeto de debate, tanto desde el punto de vista científico y filosófico como social: hay dudas sobre los mecanismos exactos de su funcionamiento y su capacidad real de comprensión del lenguaje, y sus aplicaciones plantean dilemas éticos. En este capítulo describimos cómo se ha llegado a esta tecnología y los fundamentos de su funcionamiento, permitiéndonos así comprender mejor sus capacidades y limitaciones e introducir algunos de los principales debates que rodean su desarrollo y uso.

*Palabras clave:* inteligencia artificial, procesamiento del lenguaje natural, grandes modelos de lenguaje.

## 1. INTRODUCCIÓN

Los grandes modelos de lenguaje (a menudo conocidos por las siglas en inglés *LLMs*, de *Large Language Models*) han precipitado un cambio de paradigma en el Procesamiento del Lenguaje Natural (PLN), la rama de la inteligencia artificial que busca desarrollar programas que puedan comprender y generar lenguaje humano. La clave del éxito de estos modelos es que, aprendiendo directamente a partir de texto simple, son capaces de responder adecuadamente a todo tipo de consultas, siempre que éstas sean expresables y respondibles mediante texto. Esto incluye la realización de tareas como traducción (Peng *et al.,* 2023), corrección gramatical (Fang *et al.,* 2023), resumen de textos (Pu *et al.,* 2023), respuesta a preguntas factuales (Brown *et al.,* 2020), o incluso escritura creativa (Gómez-Rodríguez y Williams, 2024), entre otras muchas. Mientras que en paradigmas anteriores (Manning y Schütze, 1999; Manning *et al.,* 2014) cada una de esas tareas requería diseñar, entrenar y ajustar un sistema específico para ella, incluyendo costosos procesos de recogida y anotación de datos especializados; los *LLMs* permiten llevarlas todas a cabo bajo el paraguas de un mismo sistema, sin ajuste específico, y con una interfaz natural que ha llevado a millones de usuarios a probar esta tecnología e integrarla en sus vidas en tiempo récord.

No obstante, para hacer un buen uso de estos modelos, es importante tener en cuenta sus limitaciones, siendo algunas de las principales las siguientes:

- Los mejores *LLMs* son sistemas computacionalmente muy costosos, lo cual está propiciando un oligopolio de facto por parte de grandes empresas tecnológicas (Bommasani *et al.,* 2021), enfrentado por modelos más pequeños (incluyendo algunos de código abierto) que por el momento no han logrado cerrar la brecha con los modelos grandes.

- Las respuestas pueden estar sesgadas o ser directamente incorrectas. Se ha llamado "alucinación" al fenómeno por el cual los *LLMs* producen respuestas que pueden estar sintácticamente bien redactadas, e incluso tener coherencia interna, pero no tienen sentido, no responden a la consulta o se alejan de la realidad.

- Se trata de sistemas opacos, donde no contamos con mecanismos fiables para determinar por qué están proporcionando una respuesta y no otra. Esto agrava el punto anterior, dado que el modelo no nos proporciona información para verificar su respuesta, haciendo desaconsejable fiarse de la salida de un *LLM* sin verificación externa.

Además, debemos ser conscientes de que, si bien algunas de estas limitaciones pueden ser fruto de la inmadurez de una tecnología incipiente y resolverse a lo largo de los próximos años, otras podrían ser intrínsecas e inevitables (Xu *et al.,* 2024; Banerjee *et al.,* 2024).

Asimismo, e incluso cuando funcionan correctamente, el uso de los *LLMs* plantea diversos retos sociales y cuestiones éticas derivadas de su uso malintencionado (por ejemplo, para generar información falsa, manipular a personas, o cometer fraude en trabajos y exámenes) o de su sobreuso (por ejemplo, sustituyendo decisiones sensibles que debería tomar una persona).

Para comprender mejor tanto las oportunidades creadas por esta tecnología como los desafíos que plantea, en este capítulo explicaremos las bases del funcionamiento de los *LLMs*, las claves técnicas de su éxito respecto a anteriores sistemas de PLN, y algunas de las más importantes preguntas abiertas y desafíos a los que se enfrenta la investigación en este campo.

Para ello, el resto del capítulo se estructura del siguiente modo: la sección 2 explica por qué los *LLMs* son tan revolucionarios, contextualizándolos mediante un resumen de los diferentes paradigmas que se han sucedido a lo largo de la historia del Procesamiento del Lenguaje Natural, para posicionar el avance que suponen. La sección 3 explica cómo funcionan los *LLMs*; y la sección 4 se basa en dicho funcionamiento para explicar lo que sabemos sobre las capacidades y limitaciones funcionales de los *LLMs*. Por último, la sección 5 resume las conclusiones del capítulo.

## 2. ¿POR QUÉ SON REVOLUCIONARIOS?

Como hemos mencionado, los grandes modelos de lenguaje (*LLMs*) han provocado un cambio de paradigma en el PLN, situando además este campo de investigación en el candelero. Pero ¿por qué son tan revolucionarios? ¿Qué es lo que los diferencia de anteriores sistemas que trabajaban en lenguaje humano? Para saberlo, conviene hacer una breve perspectiva histórica de los distintos paradigmas que se han sucedido en PLN.

El PLN tiene sus orígenes en los años 50 del siglo XX. Uno de los primeros hitos fue la propuesta del test de Turing (1950), que sugería evaluar la inteligencia de las máquinas a través de su capacidad de mantener una conversación de forma indistinguible de un ser humano. Al mismo tiempo, la Guerra Fría despierta un creciente interés por la posibilidad de traducir textos automáticamente. El experimento de Georgetown-IBM de 1954 (Hutchins, 1954) mostró al mundo un sistema capaz de traducir algunas oraciones entre inglés y ruso. Aunque muy rudimentario, los titulares grandilocuentes como "Electronic brain translates Russian"[1] difundieron entre el público general la idea de que la traducción automática era posible. Enseguida se sucedieron las predicciones optimistas, según las cuales el problema estaría resuelto definitivamente en pocos años, y la lingüística computacional (que se puede definir como la disciplina científica cuya aplicación práctica es el PLN) empezó a recibir financiación, con el consiguiente logro de avances en la tecnología. Desde estos orígenes hasta la actualidad, se podría dividir la evolución tecnológica del PLN en tres etapas, según la manera en que se ha venido "enseñando" a las máquinas a trabajar con lenguaje humano.

La primera de estas etapas abarcaría hasta finales de los años 80. En ella, los sistemas de PLN se basaban casi exclusivamente en reglas escritas a mano por expertos. Noam Chomsky (Chomsky, 1957) describe cómo sistematizar la gramática de un idioma como el inglés mediante reglas sintácticas, que permiten generar e interpretar oraciones. Basándose en este principio, se desarrollan analizadores sintácticos capaces de descomponer las oracio-nes en sus componentes gramaticales, facilitando extraer información estructurada a partir de textos. Del mismo modo, se construyen también sistemas basados en reglas para tareas como traducir textos, responder preguntas o incluso mantener una conversación, como el conocido sistema Eliza (Weizenbaum, 1966). El problema de estos enfoques es que tienen serios problemas de escalabilidad: resulta muy costoso escribir reglas manualmente para cada problema, idioma y dominio de aplicación; y es muy difícil capturar todas las variaciones y excepciones del lenguaje humano, que es cambiante y ambiguo.

---

[1] https://aclanthology.org/www.mt-archive.info/ChemEngNews-1954.pdf

Hacia finales de los 80 comienzan a aparecer, y pronto se vuelven dominantes, los sistemas basados en aprendizaje estadístico. Estos no necesitan reglas, sino que se enseña el idioma a las máquinas mediante conjuntos de ejemplos junto con algoritmos de aprendizaje automático que aprenden de ellos. Por ejemplo, para entrenar un traductor automático, se utilizaría un corpus de oraciones en el idioma origen con su traducción al idioma destino (Brown *et al.,* 1990). Este tipo de enfoques son más escalables y baratos que los basados en reglas, dado que no hace falta involucrar a expertos en el diseño del sistema mientras se cuente con datos de buena calidad (como ejemplifica la famosa frase de Fred Jelinek: "Every time I fire a linguist, the performance of the system goes up" [Hirschberg, 1998]).

A lo largo de esta etapa, la calidad de los algoritmos de aprendizaje automático va mejorando según avanzan las investigaciones, especialmente con la irrupción del llamado aprendizaje profundo (*deep learning*), una evolución que devolvió a primera línea las olvidadas redes neuronales, en la primera mitad de la década de 2010. Los sistemas con aprendizaje automático basado en *deep learning* representan las palabras en un espacio continuo de vectores densos (Mikolov *et al.,* 2013), en lugar de como entidades discretas, que se usan como entrada a redes neuronales que aprenden diferentes tareas (Cho *et al.,* 2014). Sin embargo, las mejoras cuantitativas en la precisión de los sistemas logradas por estas tecnologías no cambian la limitación fundamental de esta etapa: se trata de sistemas de propósito específico, es decir, que llevan a cabo una tarea concreta. Entrenando una red neuronal con oraciones en castellano y sus traducciones al inglés, podemos lograr un sistema de traducción automática efectivo para traducir textos de castellano a inglés. Pero será inútil para otros idiomas que no estén en el conjunto de entrenamiento, y más aún para otras tareas (como responder preguntas o resumir textos). Para ello, necesitaríamos entrenar un sistema distinto con un conjunto de datos ajustado a la tarea deseada. Así, si deseamos resumir textos en castellano, tendremos que crear otro sistema diferente, desde cero, entrenándolo con un corpus de textos en castellano con sus resúmenes, y así con cada nueva tarea que se quiera acometer.

Y es precisamente esta limitación la que viene a resolver la tecnología que motiva este capítulo, los grandes modelos de lenguaje, que han inaugurado una aún incipiente tercera etapa del desarrollo del PLN. Desde un punto de vista técnico, los grandes modelos de lenguaje provienen del escalado de los modelos neuronales de la etapa anterior (redes neuronales más grandes gracias a los avances en *hardware*, datos de entrenamiento más grandes, y arquitecturas neuronales mejores, como los Transformers [Vaswani *et al.,* 2017]). Pero desde un punto de vista más amplio, estos modelos, aun proviniendo de una evolución y no una ruptura con los anteriores, suponen un cambio de paradigma, perfectamente resumido por el título del artículo que introdujo uno de los modelos más transformativos, GPT-2 (Radford *et al.,* 2018), y que se podría perfectamente considerar el hito que inaugura la nueva etapa: *Language Models are Unsupervised Multitask Learners*. En otras palabras, los grandes modelos de lenguaje ya no necesitan datos especializados en una tarea (como oraciones junto con su traducción a otro idioma, o textos con su versión resumida) sino que aprenden a manejar el idioma a partir de cantidades gigantescas de texto, descargado tal cual de Internet u otras fuentes, sin necesidad de que algún humano los adapte a la tarea específica a realizar (*unsupervised*)... pues de hecho, no están restringidos a una tarea concreta sino que el mismo sistema puede llevar a cabo una variedad de tareas que se le pidan (*multitask*). Esto supone una auténtica revolución en la práctica, pues no solamente se obvia la necesidad de datos especializados para cada tarea, idioma y dominio de aplicación al que se quiera aplicar PLN, sino que se cuenta con sistemas de propósito general (como ChatGPT) que interactúan con el usuario final en su idioma, que puede hacer todo tipo de solicitudes sin necesidad de aprender a utilizar un *software* especializado. Por lo tanto, los grandes modelos de lenguaje hacen más versátil y universalmente accesible el PLN, a la vez que nos acercan a la superación del test de Turing.

Este cambio de paradigma, dada su reciente aparición y la vertiginosa velocidad con la que se van sucediendo los nuevos modelos y avances, suscita muchas preguntas, tanto entre los especialistas en el tema como entre el público en general. Por ejemplo, una cuestión en debate es si estos grandes modelos de lenguaje son realmente capaces de entender el lenguaje, en mayor o menor grado, o únicamente simulan respuestas coherentes sin entender en absoluto. Otras cuestiones de actualidad son si estos modelos son o pueden llegar a ser creativos, si podrían llegar a adquirir consciencia, y toda la serie de debates éticos y sociales que se plantean en torno a ellos (su fiabilidad, posibles malos usos para generar desinformación, o hasta qué punto podrían o deberían sustituir a los humanos en ciertos roles, por ejemplo).

Si bien muchas de estas cuestiones no tienen de momento respuestas definitivas que susciten consenso, para entender el debate y evitar posturas simplistas es necesario comprender algunos fundamentos de cómo funcionan los *LLMs*. Esto es lo que describimos en la siguiente sección, evitando deliberadamente el típico enfoque técnico que disecciona sus componentes (pues no es realmente necesario para este fin, haría la explicación menos accesible, y es posible que sea contingente – nada garantiza que, dentro de unos años, las arquitecturas neuronales que hacen funcionar los *LLMs* no sean radicalmente distintas de las actuales [Gu y Dao, 2024]) para centrarnos en lo esencial: *qué hacen* los *LLMs*, y qué es lo que hace que tengan las características que hemos descrito más arriba.

## 3. ¿CÓMO FUNCIONAN?

La esencia de un *LLM* es un concepto muy sencillo: son modelos que, a partir de cantidades masivas de textos que se usan como datos de entrenamiento, *predicen una continuación plausible de un texto*. Por ejemplo, si a un buen modelo de lenguaje le pasamos el texto "se me ha roto el", será capaz de continuar con una palabra coherente con el contexto (como puede ser "móvil", "jarrón" o incluso "corazón"). Para poder hacer eso, los *LLMs* se basan en los textos que han visto durante su entrenamiento.

### 3.1. Generando texto con cadenas de Markov

Para comprenderlo más en detalle, es conveniente remontarnos a una época muy anterior a los orígenes del PLN. En 1906, el matemático ruso Andrey Markov descubre un modelo para describir procesos estocásticos, conocido como cadena de Markov. Una cadena de Markov de orden k es un modelo que va atravesando una serie de estados, de tal modo que la probabilidad de cada nuevo estado depende de los k estados anteriores. Por ejemplo, una cadena de Markov podría asignar probabilidades al tiempo que hará mañana en función del tiempo de ayer y hoy, en función de una tabla como la tabla 1.

Tabla 1.

**Cadena de Markov de orden 2 para modelar el tiempo atmosférico**

| Ayer | Hoy | Probabilidad (mañana soleado) | Probabilidad (mañana nublado) | Probabilidad (mañana lluvioso) |
|---|---|---|---|---|
| Soleado | Soleado | 0.7 | 0.2 | 0.1 |
| Soleado | Nublado | 0.4 | 0.4 | 0.2 |
| Soleado | Lluvioso | 0.4 | 0.3 | 0.3 |
| Nublado | Soleado | 0.4 | 0.4 | 0.2 |
| Nublado | Nublado | 0.2 | 0.5 | 0.3 |
| Nublado | Lluvioso | 0.2 | 0.4 | 0.4 |
| Lluvioso | Soleado | 0.3 | 0.5 | 0.2 |
| Lluvioso | Nublado | 0.2 | 0.5 | 0.3 |
| Lluvioso | Lluvioso | 0.1 | 0.3 | 0.6 |

La tabla proporciona una probabilidad de que mañana haga un día soleado, nublado o lluvioso en función del tiempo que ha hecho ayer y hoy. Por ejemplo, si ayer estuvo un día nublado y hoy está lluvioso, la tabla (sexta fila) dice que hay una probabilidad 0.2 de que mañana sea un día soleado, 0.4 de nublado y 0.4 de lluvioso. Del mismo modo, podría-mos consultar las probabilidades para cualquier otra combinacion de días. En este ejemplo, siempre necesitamos dos días (ayer y hoy) para estimar las probabilidades del tiempo del día siguiente (mañana), porque se trata de una cadena de Markov de orden 2. El orden podría ser también superior o inferior. En el caso de este ejemplo, las probabilidades de la tabla son inventadas, pero en un caso real, nos basaríamos en datos históricos: por ejemplo, si en el pasado después de un día nublado y otro lluvioso vino un día soleado el 20 % de las veces, pondríamos un 0.2 en la correspondiente celda de la tabla. De este modo, las probabilidades se corresponderán a lo que cabe esperar de acuerdo con los datos históricos.

A partir de este modelo, podemos generar una secuencia estocástica, abarcando no solamente el tiempo que va a hacer mañana, sino también los días siguientes. Por ejemplo, supongamos que ayer hizo un día nublado y hoy está lluvioso. Podemos generar un tiempo plausible para mañana haciendo un sorteo aleatorio según las probabilidades de la tabla (es decir, donde haya una probabilidad 0.2 de que salga soleado, 0.4 de nublado y 0.4 de lluvioso). Supongamos que el resultado de este sorteo es un día lluvioso. Entonces, podemos repetir el proceso para pasado mañana: los dos días anteriores serían lluvioso y lluvioso, así que las probabilidades serían las de la última fila de la tabla (0.1 de día soleado, 0.3 de nublado y 0.6 de lluvioso). Este proceso podemos repetirlo indefinidamente para generar todos los días que queramos. A medida que nos alejemos en el futuro, será más difícil que la predicción del modelo se corresponda con lo que sucederá en realidad, pero al menos será una continuación plausible del tiempo registrado los dos últimos días.

El mismo concepto se puede aplicar para generar un texto en lenguaje humano, si consideramos que los estados son palabras[2] en lugar de condiciones meteorológicas. Dado un corpus de textos, las tablas de probabilidades se pueden estimar fácilmente a partir del corpus, contando las apariciones de cada posible *k*-grama (secuencia de *k* palabras) y las palabras que lo siguen.

Por ejemplo, si buscamos el bigrama (2-grama) "unidos de" en el corpus Bruno, un pequeño corpus de un millón de palabras en español con una interfaz web de libre acceso y fácil de usar para el lector que quiera experimentar (Spanish Concordancer - Bruno Spanish) Corpus; y nos fijamos en qué palabra viene después, obtendremos la tabla 2, donde vemos que este bigrama aparece casi siempre seguido de "Norteamérica" o "América" (asociado, como se puede deducir, a referencias a los EE. UU.) aunque también pueden aparecer otras palabras, como "trabajar" (en la oración "El compromiso de Estados Unidos de trabajar con los países en vías de desarrollo..."). La poca variedad de palabras se debe al pequeño tamaño del corpus: si tuviésemos más datos, sin duda podríamos observar más palabras que pueden aparecer después de "unidos de", y nuestras estimaciones de probabilidad serían también más precisas.

---

[2] En realidad, los grandes modelos de lenguaje no trabajan con palabras, sino que dividen el texto en subpalabras que típicamente se obtienen mediante un algoritmo de codificación de pares de bits (Gage, 1994). Esto se hace de este modo por diversos motivos técnicos, incluyendo una mayor eficiencia y flexibilidad para trabajar con palabras que el modelo no ha visto antes. Por simplicidad y claridad de exposición, en nuestra explicación ignoraremos este aspecto y supondremos que se trabaja con palabras, lo cual también sería una implementación posible (aunque no sea la más eficaz y eficiente) y no cambia la comprensión de los conceptos fundamentales.

Tabla 2.

**Frecuencias y probabilidades estimadas que se obtienen para modelar la palabra que viene después del bigrama "unidos de", a partir de un corpus de un millón de palabras en español**

| $w_{i-2}$ | $w_{i-1}$ | $w_i$ | Frecuencia en corpus | Probabilidad estimada |
|---|---|---|---|---|
| unidos | de | Norteamérica | 7 | 0.5385 |
| unidos | de | América | 5 | 0.3846 |
| unidos | de | trabajar | 1 | 0.0769 |

*Nota:* La notación $w_{i-2}$, $w_{i-1}$ y $w_i$ hace referencia a las palabras en las posiciones i-2, i-1 e i de un texto.

Para generar texto con nuestra cadena de Markov de orden 2, recopilaríamos datos como los de la tabla 2 para todos los bigramas del corpus (es decir, tendríamos una tabla exhaustiva con una fila para cada posible bigrama y su continuación, al modo de la tabla 1, o si se prefiere, tablas individuales como la tabla 2 para cada combinación de bigramas). A partir de esto, el proceso de generación es sencillo:

- Empezar con un k-grama cualquiera del corpus. Este *k*-grama constituirá el principio del texto generado.

- Repetir hasta que se desee:

    (a) Para las *k* últimas palabras generadas, mirar en las tablas las posibles palabras siguientes y la probabilidad de que aparezcan a continuación de ese *k*-grama.

    (b) Hacer un sorteo para elegir una de estas palabras siguientes, utilizando dichas probabilidades (es decir, cada una de las posibles palabras siguientes tendrá la probabilidad de salir que indique la tabla).

    (c) La palabra seleccionada se añade al final del texto generado.

Así, si por ejemplo el bigrama inicial fuese "las mujeres", nuestro algoritmo de generación consultaría las palabras que aparecen a continuación de ese bigrama en el conjunto de entrenamiento, obteniendo los resultados que se muestran en el bloque superior de la tabla 3. A continuación, escogería aleatoriamente una palabra de entre las posibles continuaciones, con probabilidad proporcional a las estimaciones de la tabla. En el caso de que la palabra así escogida fuese "son", el texto generado sería ahora "las mujeres son" y, en la siguiente iteración del algoritmo, se consultarían las palabras que aparecen a continuación de "mujeres son" en el corpus de entrenamiento, obteniendo los datos del bloque inferior de la tabla 3. De ahí, se esco-gería de nuevo una palabra al azar (por ejemplo, "muy"), quedando el texto como "las mujeres son muy". Este proceso se puede seguir iterando para generar un texto tan largo como se desee.

Tabla 3.

**Frecuencias y probabilidades estimadas que se obtienen para modelar la palabra que viene después de los bigramas "las mujeres" y "mujeres son", a partir de un corpus de un millón de palabras en español.**

| $w_{i-2}$ | $w_{i-1}$ | $w_i$ | Frecuencia en corpus | Probabilidad estimada |
|---|---|---|---|---|
| las | mujeres | que | 7 | 0.0414 |
| las | mujeres | son | 6 | 0.0355 |
| las | mujeres | en | 5 | 0.0296 |
| las | mujeres | no | 5 | 0.0296 |
| las | mujeres | se | 5 | 0.0296 |
| las | mujeres | ... | ... | ... |
| mujeres | son | como | 1 | 0.1667 |
| mujeres | son | mayoría | 1 | 0.1667 |
| mujeres | son | tales | 1 | 0.1667 |
| mujeres | son | un | 1 | 0.1667 |
| mujeres | son | muy | 1 | 0.1667 |
| mujeres | son | fundadas | 1 | 0.1667 |

Mediante este proceso estocástico, podemos generar texto que parece lenguaje. La tabla 4 muestra el resultado de aplicar este proceso con la Biblia como conjunto de entrenamiento[3], y el efecto de diferentes valores de *k*. En todos los textos así generados se puede apreciar el léxico y temática bíblica, dado que es la fuente de la que el modelo obtiene las palabras; pero las características sintácticas y semánticas del texto son muy diferentes según el valor de *k*. Para *k*=1, el texto carece de coherencia y tiene bastantes errores sintácticos (como "llegué a los pecados que todos"), además de no tener sentido lógico. Al aumentar el valor de *k*, el texto va cobrando coherencia, los errores sintácticos van desapareciendo (véase *k*=3, donde la sintaxis está cerca de ser correcta), y se le puede dar al menos un sentido a fragmentos del texto, aunque en su conjunto carece de sentido lógico por no tener consistencia suficiente. El aumento progresivo de coherencia y la disminución de los errores sintácticos a medida que aumenta *k* se debe a que la cadena de Markov garantiza que, por lo menos, cada (*k*+1)-grama del texto resulte plausible (dado que, por construcción, sabemos que cada (*k*+1)-grama del texto aparece alguna vez en el conjunto de entrenamiento).

---

[3] Santa Biblia: antigua versión de Casiodoro de Reina (1569), revisada por Cipriano de Valera (1602) : otras revisiones, 1862, 1909 y 1960. Revisión de 1960. Disponible en https://raw.githubusercontent.com/kblok/RNN-bible-generator/refs/heads/master/biblia.txt

Tabla 4.

**Generación de textos bíblicos mediante cadenas de Markov para distintos valores de $k$**

| Valor de $k$ | Texto generado |
|---|---|
| 1 | porque a Jehová había prevaricado con los animales del templo de Isacar, el borde de ellos le deja dormir conmigo, llegué a los pecados que todos, y recogieron doble su tiempo, de Jehová, tu vida, Y su cerrojo, se os anunciamos que Dios había agua salada con los sirvientes del Jordán, como un ángel de Parvaim. |
| 2 | Vino a mí palabra de Jehová llenó la tierra. Pero a vosotros con nuestros sacrificios y alabanzas y acción de gracias y la verdad, la cual dio a luz hijo o hija, hermano, o su hija fue sanada desde aquella hora. Y edificó allí altar a Jehová; la grosura del sacrificio de paz os santifique por completo. |
| 3 | Yo me acordaré de mi pacto con ellos, dijo Jehová: El Espíritu mío que está sobre tus lomos, y descalza las sandalias de tus pies. Y él puso su diestra sobre mí, diciéndome: No temas; yo soy el que los hirió, sino que se levantó en el campo de Joab está junto al mar de Galilea, y les enseñaba en los días de adversidad. |
| 10 | Y las aguas del mar faltarán, y el río se agotará y secará. Y se alejarán los ríos, se agotarán y secarán las corrientes de los fosos; la caña y el carrizo serán cortados. La pradera de junto al río, de junto a la ribera del río, y toda sementera del río, se secarán, se perderán, y no serán más. |

Podría pensarse entonces que, para alcanzar un texto totalmente coherente, bastaría con aumentar $k$ hasta donde sea necesario. Sin embargo, si hacemos eso, nos encontraremos con otro problema: el agotamiento del espacio muestral. Si observamos el texto generado para $k$=10 en la tabla 4 podremos ver que, efectivamente, es muy coherente. El problema es que se trata de una copia literal de un fragmento del conjunto de entrenamiento, dado que éste no es lo suficientemente grande como para que encontremos en él varias muestras de un 10-grama. Por lo tanto, dado un 10-grama (como puede ser "Y las aguas del mar faltarán, y el río se"), nuestra tabla de posibles continuaciones solamente tendrá una opción ("agotará"), con probabilidad 1, y lo que estará haciendo el modelo estocástico es reproducir el texto bíblico en lugar de generar texto nuevo.

La generación de textos con cadenas de Markov, pues, está muy lejos de la apariencia de inteligencia que se atribuye a los textos generados por *LLMs*, y es de dudosa aplicabilidad práctica por las limitaciones que tiene: con $k$ pequeño, los textos no tienen coherencia, y con $k$ más grande, enseguida nos encontramos con la barrera de la escasez de datos. Se podría pensar en mitigar esto último con conjuntos de entrenamiento más grandes; pero esto por sí solo tiene un alcance muy limitado: dado que la probabilidad de encontrar un determinado $k$-grama en un texto disminuye de manera exponencial respecto a $k$, ni aunque juntásemos todos los textos escritos a lo largo de la historia de la Humanidad podríamos tener estimaciones de probabilidad aceptables para $k$=10. Basta pensar que, si elegimos una secuencia de $k$ palabras consecutivas de este texto o cualquier otro que hayamos escrito, lo más probable es que no haya sido escrita nunca antes (salvo en casos concretos de texto formulaico, como frases hechas o textos legales). Por lo tanto, por muy grande que sea nuestro conjunto de entrenamiento, es casi seguro que un modelo con *k*=10 se va a limitar a copiar literalmente.

Sin embargo, estas limitaciones de las cadenas de Markov se pueden atajar si nos salimos de las reglas estrictas que nos impone el modelo, haciéndolo más flexible. Desde el descubrimiento de las cadenas de Markov fueron apareciendo algunas mejoras incrementales en este sentido, como las técnicas de suavizado (Chen y Goodman, 1996), interpolación (Jelinek, 1980) o *back-off* (Katz, 1987). Sin embargo, el avance que realmente permitiría desbloquear modelos generativos de lenguaje con valores grandes de k fueron los *modelos de lenguaje neuronales*.

### 3.2. De las cadenas de Markov a los modelos neuronales

Los modelos de lenguaje neuronales son, como las cadenas de Markov, modelos que generan una continuación plausible de un texto a partir de lo que han observado en el conjunto de entrenamiento. Sin embargo, en lugar de basarse puramente en cálculos de probabilidades condicionales, la generación se produce mediante redes neuronales. Esto hace que estos modelos puedan generar textos de manera más flexible, soportando así longitudes de contexto (*k*) más largas. Si bien explicar en detalle el funcionamiento de estos modelos escapa a los objetivos de este capítulo, sí cabe destacar que son principalmente tres los avances que han posibilitado lograr esta flexibilidad de la que las cadenas de Markov carecían: las representaciones densas del lenguaje, los avances técnicos en redes neuronales, y la disponibilidad de enormes conjuntos de datos obtenidos de Internet. A continuación se describe brevemente cada uno de estos avances y su relevancia.

#### *Representaciones densas del lenguaje*

Las representaciones densas del lenguaje, conocidas en inglés como *word embeddings*, son vectores numéricos que capturan el significado de las palabras al ubicarlas en un espacio vectorial donde palabras con significados similares están cerca unas de otras. Típicamente, el vector que representa cada palabra está formado por unos pocos cientos de componentes en coma flotante (aproximación informática de los números reales). La popularización de estas representaciones se produjo, sobre todo, a raíz del modelo word2vec de Mikolov *et al.* (2013), que obtiene vectores para las palabras a partir de una sencilla red neuronal. Desde entonces, han ido apareciendo diferentes alternativas (Pennington *et al.,* 2014), incluyendo representaciones contextualizadas, donde el vector de cada palabra varía según el contexto en que aparezca (Devlin *et al.,* 2019).

El uso de representaciones densas del lenguaje tiene varias ventajas sobre considerar cada palabra una entidad discreta. Una de ellas es que las redes neuronales se adaptan mejor a trabajar con vectores densos de números reales que con entidades discretas. Pero la principal ventaja es, sobre todo, la flexibilidad que añaden al modelar de forma natural la similitud entre palabras. En las cadenas de Markov vistas en el apartado 3.1, así como en otros modelos estadísticos tradicionales usados en PLN, cada palabra se representa como una entidad atómica sin relación aparente con las demás: un modelo de este tipo verá las palabras "abanico", "avión" y "aeroplano" como cadenas de texto diferentes que se traducen interiormente a algún tipo de identificador numérico; pero el modelo no tendrá constancia de que las dos últimas se parecen mucho más entre sí que a la primera. Así, si en una cadena de Markov queremos continuar el texto "perdió el control del aeroplano" y ese $k$-grama exacto no aparece en el conjunto de entrenamiento, pero sí aparece "perdió el control del avión", no podremos aprovechar esta similitud para generar nuestra continuación del texto. En cambio, un modelo neuronal sí podrá hacerlo: en lugar de buscar coincidencias exactas de palabras en una tabla, el modelo neuronal trabaja en un espacio continuo de representaciones vectoriales, así que puede aprovecharse de fragmentos de texto similares, pero no iguales, que haya visto en el entrenamiento. Esto le proporciona flexibilidad adicional frente a la rigidez de una cadena de Markov, y la capacidad de sacar mucho más partido a los datos.

*Avances técnicos en redes neuronales*

Las redes neuronales son modelos computacionales inicialmente inspirados en la estructura y funcionamiento del cerebro humano, diseñados para aprender la relación entre entradas y salidas deseadas para problemas complejos. Están formadas por unidades de procesamiento llamadas "neuronas", organizadas en capas que se conectan entre sí. En una red neuronal, las neuronas reciben entradas, realizan cálculos y transmiten los resultados a otras neuronas en la siguiente capa a través de conexiones que las unen. Mediante un proceso de entrenamiento, estas conexiones se ajustan usando algoritmos de optimización, permitiendo que la red aprenda patrones a partir de datos y mejore en su tarea específica, como puede ser el procesamiento de textos.

Si bien las redes neuronales son conocidas desde mediados del siglo XX, a finales de siglo era una tecnología bastante olvidada pues, con lo que se sabía hasta entonces, no solían ser la mejor opción en la práctica para la mayoría de problemas, siendo claramente superadas por otras alternativas como las máquinas de vectores soporte. De hecho, como señalan LeCun *et al.* (2015), las redes neuronales fueron "en gran medida abandonadas por la comunidad de aprendizaje automático" porque "se pensaba ampliamente que [la forma en la que aprendían a partir de los datos] era inviable". Su resurgir es un ejemplo de cómo la inversión en investigación básica, incluso sin aplicaciones claras a la vista, es clave para el avance científico: durante la década de 2010, se logran importantes mejoras en las redes neuronales con una serie de descubrimientos que posibilitan el llamado aprendizaje profundo (Goodfellow *et al.*, 2016), un enfoque que transforma el campo de la inteligencia artificial. Gracias a avances en algoritmos, como el uso de capas de redes neuronales más profundas y la implementación de técnicas de regularización y optimización, así como el aumento de la capacidad de cómputo facilitado por el uso de *GPUs*, estas redes comienzan a alcanzar un rendimiento notable en tareas complejas, incluyendo en procesamiento del lenguaje natural (Goldberg y Hirst, 2017).

Si bien inicialmente las redes neuronales usadas en PLN son arquitecturas ya conocidas de otros problemas que se adaptan para procesar lenguaje (Goldberg y Hirst, 2017), el avance que supondría el pistoletazo de salida para los modelos de lenguaje neuronales que más tarde evolucionarían a grandes modelos de lenguaje es el desarrollo de una arquitectura neuronal específicamente pensada para el procesado de textos: los *Transformers* (Vaswani *et al.,* 2017). Esta arquitectura cuenta con un mecanismo de "autoatención" (*self-attention*) que permite relacionar directamente cada elemento de una secuencia de datos (como una palabra en una oración) con todos los demás, capturando así las dependencias a larga distancia que caracterizan al lenguaje humano. Los modelos basados en *Transformers* enseguida empezaron a mostrar una especial eficacia en todo tipo de tareas de PLN, superando a las arquitecturas neuronales anteriores (Devlin *et al.,* 2019), y están en el núcleo de todos los grandes modelos de lenguaje más conocidos; aunque también estén empezando a explorarse otras arquitecturas alternativas que podrían sustituirlos (Gu *et al.,* 2024).

### *Grandes volúmenes de datos*

El tercer avance clave para lograr modelos más eficaces para continuar de modo plausible un texto es el entrenamiento con cantidades cada vez mayores de datos. Si bien comentamos en el apartado 3.1 que esto por sí solo no sería suficiente para romper las limitaciones de las cadenas de Markov, aunque tuviésemos todo el texto que jamás se ha escrito; este factor sí que ha resultado clave en combinación con las representaciones vectoriales del lenguaje y las mejoras en las arquitecturas neuronales, que permiten precisamente explotar mejor los datos. Concretamente, muchos grandes modelos de lenguaje actuales utilizan conjuntos de entrenamiento de terabytes (Liu *et al.,* 2024), que contienen billones españoles (o trillones americanos) de palabras. Por ejemplo, LLaMa 3, un modelo de lenguaje de Meta, se entrenó sobre 15 billones españoles de "tokens" (fragmentos de palabras) (Dubey *et al.,* 2024), lo cual puede suponer entre 7 y 10 billones españoles de palabras. Para hacerse una idea de lo enorme que es esta cifra, se ha estimado que todos los libros existentes en el mundo en la actualidad podrían sumar unos 16 billones españoles de palabras[4].

### 3.3. De la predicción de palabras a la realización de tareas

Con los tres ingredientes explicados en la sección anterior, podemos conseguir modelos neuronales de lenguaje que hacen lo mismo que las cadenas de Markov: generar una continuación plausible de un texto, a partir de lo que han visto en su conjunto de entrenamiento, de acuerdo con las probabilidades que se deducen de los datos (continuaciones más típi-cas o comunes serán más probables, mientras que las más atípicas serán más improbables). Sin embargo, podemos hacerlo sin las limitaciones intrínsecas de las cadenas de Markov: los modelos de lenguaje neuronales nos permiten usar tamaños de contexto de cientos de palabras sin que aparezca el problema de la escasez de datos, mejorando en gran medida la coherencia de los textos generados.

---

[4] https://www.educatingsilicon.com/2024/05/09/how-much-llm-training-data-is-there-in-the-limit/#all-books

Así, en la tabla 5 se muestran tres textos generados con una versión en español de GPT-2[5], un *LLM* pequeño para los estándares actuales, entrenado sobre 11.5 GB de datos. Los textos fueron generados proporcionando la entrada *La torre de Hércules es,* y permitiendo que el modelo predijese palabras a partir de ahí. Como se puede apreciar, si nuestro objetivo es generar textos en correcto español, lo hemos cumplido con creces: los textos generados son sintácticamente correctos, e incluso semánticamente coherentes. Ahora bien, el contenido de los textos no se ajusta a la realidad: en cada caso se genera una continuación aleatoria que sitúa la torre de Hércules en un lugar diferente, y nunca en su ubicación real, A Coruña, además de proporcionar otros datos que tampoco son veraces. Por supuesto, esto no debería sorprendernos: como hemos reiterado varias veces a lo largo de este capítulo, estos modelos están entrenados simple y exclusivamente para *generar una continuación plausible del texto.* En ningún momento de su proceso de entrenamiento se hace nada para procurar que generen textos factualmente ciertos.

Tabla 5.

**Generación de textos con GPT-2 a partir de la cadena "La torre de Hércules es"**

| |
|---|
| *La Torre de Hércules es* una de las construcciones más antiguas de la ciudad de Alicante. Se encuentra en la Plaza de la Constitución, en el centro de la ciudad. Fue construida en el siglo XVI y es una de las más antiguas de la ciudad… |
| *La Torre de Hércules es* una torre de piedra, situada en la ciudad de Palma de Mallorca, a una altura de 86 metros. La torre de Hércules data del siglo XIV. Se construyó siguiendo un modelo inspirado en los edificios de Córcega… |
| *La Torre de Hércules es* la construcción del siglo XI más antigua de la que se conserva en Cataluña, y su estructura original ha sido reconstruida en múltiples ocasiones, pero únicamente la tercera y última ha permanecido en pie… |

Lo realmente sorprendente, y que supuso un giro inesperado para la comunidad de investigación en PLN (Wei *et al.,* 2022), es que simplemente escalando el modelo (entrenándolo con más datos, y haciendo la propia red neuronal más grande, con más neuronas y conexiones), sin ninguna mejora cualitativa en especial, se consigue que empiece a decir la verdad. La tabla 6 muestra varias generaciones de texto a partir del mismo principio que la tabla 5; pero esta vez provienen de BLOOM (Workshop *et al.,* 2023), un modelo más grande y entrenado con unos 200.000 millones de palabras, pero que por lo demás no es esencialmente distinto de GPT-2. Como se puede ver en la tabla, el resultado son textos que ya no solo son sintácticamente correctos y semánticamente coherentes, sino que además dan en su mayoría datos ciertos (aunque no con fiabilidad total: por ejemplo, no es cierto que la Torre de Hércules esté en un islote).

---

[5] https://huggingface.co/DeepESP/gpt2-spanish

Tabla 6.

**Generación de textos con BLOOM a partir de la cadena "La torre de Hércules es"**

| |
|---|
| *La Torre de Hércules es* el faro más antiguo del mundo en funcionamiento y el único del mundo romano que sigue dando servicio. La torre se encuentra situada en la península de la ciudad de A Coruña y está construida sobre una base rocosa de 40 metros de altura... |
| *La Torre de Hércules es*, sin lugar a duda, uno de los símbolos de la ciudad de La Coruña y uno de los faros en funcionamiento más antiguos del mundo. La historia del faro empieza en el siglo I d.C. cuando los romanos lo construyeron sobre las ruinas de un faro etrusco... |
| *La Torre de Hércules es* el faro más antiguo del mundo todavía en funcionamiento, se encuentra situado en la ciudad de A Coruña, Galicia, sobre un islote en la península de la Torre. El faro data del siglo I.... |

Por otra parte, si utilizamos el mismo modelo para continuar una pregunta *(¿Qué es la Torre de Hércules?)*, obtendremos los textos de la tabla 7. Como vemos, el resultado es que, a veces, el modelo responde la pregunta, y otras veces no. Este comportamiento es lógico: dado un texto que contiene una pregunta, una continuación plausible del texto es responder dicha pregunta (a menudo tras una pregunta aparece su respuesta, y el sistema habrá visto este patrón en el conjunto de entrenamiento) pero también hay otras continuaciones plausibles que no implican responder (por ejemplo, en el texto de un examen pueden aparecer preguntas seguidas de más preguntas, y sin ninguna respuesta).

Tabla 7.

**Generación de textos con BLOOM a partir de la cadena "¿Qué es la Torre de Hércules?"**

| |
|---|
| *¿Qué es la Torre de Hércules?* La Torre de Hércules (A Coruña) es el faro en funcionamiento más antiguo del mundo. Símbolo de la ciudad... |
| *¿Qué es la Torre de Hércules?* La Torre de Hércules es uno de los monumentos más emblemáticos de España, es un faro construido por los romanos en el Siglo I... |
| *¿Qué es la Torre de Hércules?* ¿Cuánto mide? ¿Qué altura tiene la Torre de Hércules? ¿Dónde está? ¿En qué lugar se ubica? ¿Cuánto tiempo es necesario para subir? ... |

Por lo tanto, mediante el escalado de los modelos de lenguaje neuronales, hemos obtenido un sistema que es capaz de proporcionar información veraz, aunque sin mucha fiabilidad, y que incluso responde preguntas, aunque solamente a veces. Más allá de la curiosidad de generar textos al azar, el modelo podría ser realmente útil, si lográramos dotarlo de consis-tencia: por ejemplo, nos gustaría que ante una pregunta, siempre intentarse responder, como en los dos primeros ejemplos de la tabla 7, en lugar de generar otras preguntas como en el último. Dicho de otro modo, no solo queremos que el modelo genere una continuación plausible del texto (y a ser posible, veraz); sino que dentro de las posibles continuaciones plausibles, queremos potenciar algunas (como responder preguntas) y desincentivar otras (como hacer más preguntas).

Esto último se consigue con las técnicas denominadas de "ajuste de instrucciones" (*instruction tuning*), que se utilizan para ajustar el modelo de modo que "prefiera" generar el tipo de continuaciones del texto que los humanos queremos. Existe una variedad de técnicas de instruction tuning, pero sobre todo se dividen en dos tipos, que se suelen aplicar en conjunto (Ouyang *et al.,* 2022):

- Ajuste supervisado: se proporcionan al modelo ejemplos del tipo de respuestas que queremos, siguiendo un formato instrucción-respuesta para que el modelo lo imite.

- Aprendizaje por refuerzo con realimentación humana (en inglés, *reinforcement learning with human feedback* o *RLHF*): se proporciona al modelo realimentación humana sobre la calidad de sus respuestas, para que prefiera las respuestas más deseables y se aleje de las menos deseables. Por ejemplo, dadas las respuestas de la tabla 7, generadas por el propio modelo, un evaluador humano marcaría las dos primeras como deseables y la tercera como indeseable. Estos datos se usarían para ajustar el modelo, haciendo que prefiera responder a las preguntas.

Es con estas técnicas cómo se obtienen los *LLMs* tal como los conocen los usuarios finales. Sistemas como ChatGPT o Claude, pues, no son más que modelos neuronales de lenguaje entrenados para proporcionar una continuación plausible de un texto (funcionalmente lo mismo que las cadenas de Markov, pero con mejor desempeño gracias a las tecnologías neuronales y a la mayor capacidad de cómputo y datos) que se someten a un proceso de ajuste para que, dentro de las posibles continuaciones plausibles, prefieran aquéllas que nosotros indicamos como prioritarias.

### 4. ¿QUÉ CAPACIDADES Y LIMITACIONES TIENEN?

Como hemos visto en la sección anterior, los *LLMs* no están entrenados explícitamente para responder a las peticiones del usuario, ni siquiera para que sus textos reflejen la realidad, sino solamente para generar continuaciones plausibles de un texto a partir de lo observado en los datos de entrenamiento. En este sentido, los modelos pequeños se limitan a generar texto coherente, pero "inventado", que no refleja la realidad ni responde a lo que pide el usuario. Sin embargo, cuando los modelos se hacen más grandes (con más neuronas y conexiones en su red) y se entrenan con mayores volúmenes de datos, espontáneamente el contenido de sus textos se va volviendo más fiable, a la vez que surgen habilidades como la de responder preguntas. A este respecto, es importante resaltar que el ajuste de instrucciones sirve para dar más consistencia a los modelos y hacerlos más previsibles (por ejemplo, como hemos visto, para que *siempre* respondan a las preguntas que se les hacen, en lugar de solo a veces), y también para mitigar posibles sesgos causados por los datos de entrenamiento (los modelos pueden producir respuestas con sesgos indeseables como machismo o racismo si éstos están presentes en los datos de entrenamiento, como suele ser el caso en conjuntos de datos no filtrados descargados de Internet); pero en ningún caso sirve para añadir capacidades. En los ejemplos vistos en la sección anterior (tabla 7), se puede ver cómo el modelo BLOOM en bruto, sin ningún ajuste, ya es capaz de responder preguntas. El ajuste de instrucciones solo hace que esa capacidad se aproveche de forma más consistente para el usuario final.

Estas nociones sobre el funcionamiento de los *LLMs* nos permiten explicar lo que probablemente sea su mayor limitación tecnológica: el conocido fenómeno de las alucinaciones (Ji *et al.,* 2023), que es como se ha dado en llamar a las situaciones en las que un *LLM* genera texto que suena plausible, pero contiene información factualmente incorrecta o incluso sin sentido. Conociendo el funcionamiento de los *LLMs* podemos deducir que, desde un punto de vista técnico, las alucinaciones no son un fallo del sistema (entendiendo fallo como comportamiento anómalo o no previsto); sino que son una consecuencia directa del funcionamiento normal del sistema: entrenamos modelos para generar texto plausible, y eso es lo que nos dan. Las alucinaciones son especialmente prevalentes cuando el modelo no es capaz de proporcionar una respuesta correcta (por ejemplo, porque se le pregunta por algo que no tiene en su conjunto de entrenamiento). De hecho, Hicks *et al.* (2024) argumentan que las alucinaciones se entienden mejor considerándolas como *bullshit*, en el sentido descrito por Frankfurt en su libro On Bullshit (Frankfurt, 2009). Para Frankfurt, "bullshit" es una forma de comunicación en la que el emisor no se preocupa por la verdad ni la falsedad de lo que dice, sino únicamente por el efecto que sus palabras pueden tener. Según Hicks *et al.* (2024), los grandes modelos de lenguaje serían generadores de *soft bullshit*, que es aquélla que no se emite con intención de manipular. Haciendo un símil humano, sería el tipo de discurso que tienen las personas que no quieren admitir que no saben de un tema, y dicen cualquier cosa para salir del paso cuando se les pregunta sobre él.

Por supuesto, los investigadores en PLN y creadores de *LLMs* están trabajando para reducir las alucinaciones al mínimo posible. El ajuste de instrucciones (Ouyang *et al.,* 2022), visto en la sección anterior, juega un papel importante en ello, pues nos proporciona una forma de incentivar al modelo para que priorice la generación de textos ciertos por encima de los falsos. Otra línea muy activa de investigación es la generación aumentada con recuperación (en inglés, *retrieval-augmented generation* [Lewis *et al.,* 2020]), una técnica en la que se hace que el modelo consulte una base de datos o fuente de información externa para recuperar información relevante antes de generar una respuesta. Esto no solamente permite que el modelo combine su conocimiento interno con información actualizada o específica del contexto, mejorando la precisión de sus respuestas, sino que reduce las alucinaciones porque minimiza las situaciones en las que el modelo no tiene acceso a la respuesta, incurriendo en la soft bullshit que antes mencionábamos. Sin embargo, al usar *LLMs* debemos ser conscientes de que, si bien estas técnicas reducen las alucinaciones lo suficiente para proporcionarnos modelos útiles y aplicables en multitud de situaciones, ninguna de ellas garantiza eliminarlas por completo. Las alucinaciones, como hemos explicado, son una característica intrínseca de la manera en que están diseñados y entrenados los *LLMs*, y han sido descritas como una "limitación innata" e "inevitable" (Xu *et al.*, 2024) y que va a existir "siempre" (Banerjee *et al.*, 2024). Por lo tanto, al usar LLMs, es importante no otorgar nunca presunción de veracidad a la información que proporcionan, y verificarla por otras fuentes si es importante que sea cierta.

Volviendo a los aspectos positivos de los *LLMs*, lo cierto es que; aunque las alucinaciones existan, en muchos casos estos modelos sí son capaces de proporcionarnos respuestas veraces, hasta el punto de que los mejores *LLMs* alcanzan unas tasas de fiabilidad muy altas siempre que las peticiones que les hacemos no requieran razonamientos complejos, saberes muy especializados, conocimiento de noticias de actualidad, o se enfoquen en algún punto débil de estos modelos, como puede ser la aritmética (los *LLMs* son modelos de lenguaje enfocados a palabras y no tienen acceso directo a qué cifras conforman un número, de ahí que no sean sorprendentes sus fracasos en este aspecto (Gambardella *et al.*, 2024): es pedirles trabajar con información a la que no tienen acceso más que indirectamente, como una persona ciega respondiendo preguntas sobre colores). El hecho es que la cantidad de cosas que *sí* pueden hacer, de manera bastante fiable, es enorme: no solamente demuestran la capacidad de responder preguntas, sino que son capaces de traducir textos, resumirlos, corregirlos, programar en distintos lenguajes de programación, e incluso tienen un buen nivel en escritura creativa.

A modo de inciso, al respecto de esto último, conviene desterrar un mito muy difundido, según el cual los *LLMs* nunca podrían generar textos creativos porque "predicen siempre la palabra más probable" o, como dice Ted Chiang en The New Yorker (Chiang, 2023) hacen "un promedio de las elecciones que otros escritores han hecho, representadas por textos encontrados en Internet; ese promedio equivale a las elecciones menos interesantes posibles, por lo que el texto generado por IA suele ser realmente insulso". Estas descripciones son simplificaciones engañosas que no reflejan el funcionamiento real de los *LLMs*: como hemos visto, éstos funcionan prediciendo la siguiente palabra, pero no es necesariamente ni la palabra más probable, ni un "promedio" que elimine la diversidad o lo atípico. Del mismo modo que las cadenas de Markov, lo que utilizan es una distribución estadística, donde las palabras más probables según los datos de entrenamiento se predecirán más a menudo, y las más atípicas menos a menudo. Pero nada impide que un *LLM* se desmarque con una elección realmente inusual, y de hecho esto incluso se puede fomentar, ajustando un parámetro llamado "temperatura" (Peeperkorn *et al.*, 2024) para aplanar la distribución estadística, disminuyendo la probabilidad de las palabras más probables y aumentando la de las que lo son menos. Y al margen de estas consideraciones, lo cierto es que ya hay resultados que muestran que los *LLMs* pueden generar historias que jueces humanos consideran incluso mejores que las escritas por personas (Gómez-Rodríguez y Williams, 2023). Aunque el resultado depende mucho de factores como longitud de los textos, género literario, idioma empleado, nivel de los escritores humanos con los que se compara y otras condiciones de la tarea, que hace que los resultados de este tipo de experimentos sean diversos (Marco *et al.*, 2024; Chakrabarty *et al.*, 2024); lo cierto es que no hay argumentos teóricos ni empíricos para defender que los textos generados por *LLMs* tengan por qué ser insulsos.

Pero regresando a las habilidades de los *LLMs*, la gran pregunta es: ¿por qué y cómo estos modelos las desarrollan? Lo cierto es que, a día de hoy, no lo sabemos en absoluto. Como se comentó en el apartado 3.3, el hecho de que al hacerse más grandes los modelos de lenguaje dejasen simplemente de generar textos al azar, para proporcionar respuestas con sentido, no fue algo buscado, sino un giro inesperado por los propios investigadores que lo hicieron posible. Se trata de *habilidades emergentes*: capacidades que surgen de manera aparentemente espontánea cuando el sistema alcanza una determinada escala, sin haber sido explícitamente programado o diseñado para ello. Aun hoy, aunque por supuesto hay muchos estudios intentando explicar cómo y por qué surgen estas capacidades (Schaeffer *et al.,* 2024; Du *et al.,* 2024), estamos muy lejos de lograr responder la pregunta. Y esta ignorancia sobre cómo los *LLMs* logran hacer lo que hacen implica, asimismo, una falta de transparencia: hoy por hoy, no podemos explicar por qué un *LLM* está proporcionando una respuesta y no otra, haciendo desaconsejable utilizarlos para cualquier toma de decisiones relevantes, salvo como asistentes que informen a la persona que tome la decisión final.

De hecho, tan poco sabemos hoy por hoy de los *LLMs* que incluso resulta polémico un aspecto que parece que debería ser básico: ¿los *LLMs* entienden (en algún grado) el lenguaje humano? Muchos expertos, como Bender *et al.* (2021), defienden que no lo comprenden en absoluto, basándose sobre todo en su propio funcionamiento. Como hemos visto, los grandes modelos de lenguaje generan palabras basándose en un modelo estocástico, similar en esencia a una cadena de Markov. Son, pues, lo que Bender et al. llaman "loros estocásticos", limitándose a repetir patrones lingüísticos previamente observados en los datos de entrenamiento, sin entender realmente el significado o el contexto en el que se usan. Cuando escogen una palabra dada, no lo hacen basándose en su significado, sino en mera estadística (eligiéndola al azar de una distribución). Tampoco tienen consciencia ni intenciones, con lo cual sus respuestas son lenguaje "falso", vacío de pensamiento o intencionalidad. En un experimento mental (Bender, 2020) que recuerda al clásico argumento de la habitación china de Searle (1985), Bender compara el aprendizaje de los *LLMs* con una persona que, sin saber tailandés ni conocer siquiera su alfabeto, estuviese encerrada en una biblioteca llena de libros en tailandés. El argumento es que, aun disponiendo tiempo ilimitado, sería imposible que esa persona lograra *comprender* realmente la lengua tailandesa: podría tal vez, a partir de patrones observados en los libros, ser capaz de dar respuestas convincentes a oraciones escritas en ese idioma; pero sería un mero ejercicio de reconocimiento de patrones sin comprensión real del lenguaje. Nunca podría deducir lo que significan las palabras porque no tendría referentes del mundo real para poder vincularlas a conceptos y experiencias.

Sin embargo, otros autores no están de acuerdo, y sí ven comprensión real del lenguaje (aun cuando sea diferente a la humana, y pueda tener sus limitaciones) en los *LLMs*. Aunque no sepamos muy bien cómo lo hacen, se ha visto que los *LLMs* más grandes son capaces de responder preguntas que necesitan razonar sobre objetos y propiedades del mundo físico que nunca han observado (Li *et al.,* 2021; Patel y Pavlick, 2022). En la tabla 8 se puede ver una respuesta de ChatGPT ante una petición que, para un ser humano, implicaría hacerse un modelo mental del mundo y pensar en cómo va cambiando el estado de los distintos objetos a medida que suceden los hechos narrados. Como se puede ver, el *LLM* proporciona la respuesta correcta. Aunque esto no es así el 100 % de las veces, los mejores modelos suelen tener un grado de fiabilidad bastante alto en preguntas como ésta o incluso más complejas, lo cual parece difícil de conseguir sin una cierta capacidad de entender el mundo y, por ende, el significado de las palabras.

Tabla 8.

**ChatGPT (con GPT-4) respondiendo una pregunta que parece requerir ciertas capacidades de modelado de mundo**

*Petición:* Estoy en mi habitación. Cojo un vaso lleno de agua, y le pongo un cubito de hielo. Meto el vaso dentro de una caja negra. Llevo la caja negra al salón. Saco el vaso de la caja, vierto su contenido al suelo. Vuelvo a meter el vaso en la caja. Voy a la cocina, y pongo la caja en la encimera. ¿Dónde está el cubito de hielo ahora? ¿Dónde estará dentro de veinte minutos?

*Respuesta:* Dado que viertes el contenido del vaso al suelo, el cubito de hielo, asumiendo que se encontraba en el vaso, ahora estaría en el suelo del salón. Dentro de veinte minutos, dependiendo de la temperatura de la habitación, es probable que el cubito de hielo se haya derretido, por lo que estaría en forma de agua en el suelo del salón.

Si bien se podría argumentar que tal vez los *LLMs* sean capaces de responder este tipo de preguntas mediante razonamientos superficiales o reconocimiento de patrones, sin un verdadero modelo interno del mundo, hay un experimento que proporciona pruebas sólidas de su capacidad de modelado, al menos en un entorno de juguete: Li *et al.* (2023) entrenan un modelo GPT-2 sobre listas de jugadas de un juego de mesa, Othello. El entrenamiento se lleva a cabo desde cero, es decir, el modelo no ha tenido acceso a ningún otro tipo de información en absoluto, ni siquiera ha estado expuesto a lenguaje humano. Lo único que ha visto son listas de jugadas (como "C3" o "D4") procedentes de partidas. En principio, del mismo modo que la persona encerrada en la biblioteca tailandesa no tiene información que le permita vincular las palabras que aparecen en los libros con objetos o conceptos; este modelo no tiene información que le permita saber que las listas que recibe se corresponden con un juego de tablero, cómo es dicho tablero o el número de jugadores. Sin embargo, el experimento de Li *et al.* muestra cómo el sistema, solamente a partir de las listas de jugadas, es capaz de jugar correctamente (sugiriendo jugadas legales más del 99 % de las veces) y generar una representación interna del tablero (64 neuronas que representan cada una de las casillas del tablero 8x8 del juego, cosa que los autores pudieron comprobar manipulando los bits de esta representación interna y comprobando su efecto en la partida).

Parece, pues que, de alguna manera, los grandes modelos de lenguaje son capaces de inferir *significados* a partir de las cadenas de texto que reciben: aunque, como el prisionero de la biblioteca tailandesa, no dispongan de información sobre la correspondencia entre esas cadenas y entidades reales; consiguen hacerse una idea del funcionamiento del mundo exterior (que en el trabajo de [Li *et al.*, 2023], sería el juego de Othello), a partir de dichas cadenas. Y todo esto, a pesar de que en ningún momento hemos dejado de hablar de modelos entrenados, simple y exclusivamente, para predecir la continuación plausible de un texto (o, en el caso de Othello, de una lista de jugadas). El modelo de Li *et al.* construye un modelo del tablero de Othello porque *la mejor forma de predecir la siguiente jugada de una partida es comprender el juego*; así que con este único objetivo, es capaz de adquirir comprensión (un modelo) de cómo funciona Othello. Del mismo modo, se podría hipotetizar que la mejor forma de predecir la siguiente palabra en un texto es comprender el mundo al que hace referencia el texto, y de ahí podría provenir la capacidad de modelado de mundo de los *LLMs*; aunque la manera en la que la logran siga siendo un misterio.

En cualquier caso, las obvias lagunas que aún existen en nuestro entendimiento de cómo los *LLMs* adquieren sus habilidades hacen que el debate continúe abierto sobre la cuestión de si comprenden o no el lenguaje: ni tenemos una refutación concluyente de la posibilidad de que un "loro estocástico" lo suficientemente grande y complejo no pueda adquirir una comprensión del lenguaje, ni tenemos pruebas concluyentes de que sea así, dado que el de Othello no deja de ser un ejemplo de juguete en comparación con la complejidad de comprender el lenguaje humano.

De hecho, el debate que suscita esta cuestión ha avivado, a su vez, el debate sobre qué significa exactamente "comprender" el lenguaje humano, algo en lo que estamos lejos del consenso (Søgaard, 2022; Havlík, 2023). Y es que la aparición de estos sistemas que son capaces de manejar el lenguaje de maneras que hasta ahora eran exclusivas de los humanos plantea profundas cuestiones relacionadas con el antropomorfismo. Es frecuente caer en el error de describir a los *LLMs* utilizando términos antropomórficos (Abercrombie, *et al.,* 2023) y atribuirles cualidades o intenciones humanas, ya sea para alabar su funcionamiento (por ejemplo, atribuyéndoles empatía o emociones) o para criticarlo (por ejemplo, acusándolos de mentir o engañar al usuario). No obstante, también es cuestionable descartar por completo que una entidad no humana y no consciente pueda ser el sujeto de verbos tradicionalmente exclusivos de los humanos, como "razonar", si en la práctica estos modelos logran resultados que a menudo resultan indistinguibles de los que obtendría un ser humano al realizar esas mismas acciones (Huang y Chang, 2023). Por otra parte, tampoco podemos descartar que algunas de las barreras que actualmente existen entre la inteligencia humana y las capacidades de los *LLMs* se vayan difuminando cada vez más a medida que los avances técnicos continúen mejorando estos últimos. Por ejemplo, Chalmers (2024) advierte de que aunque es improbable que los modelos de lenguaje actuales sean conscientes, debemos tomar en serio la posibilidad de que sus sucesores pudiesen llegar a serlo en un futuro.

## 5. CONCLUSIÓN

Tras explicar por qué los *LLMs* son radicalmente diferentes a anteriores tecnologías del lenguaje (sección 2), la descripción de cómo funcionan (sección 3) nos ha permitido comprender mejor sus principales capacidades y limitaciones (sección 4). Al ser esencialmente sistemas que predicen una continuación plausible de un texto a partir de distribuciones estadísticas obtenidas de los datos de entrenamiento, están sujetos a problemas como la presencia de sesgos procedentes de dichos datos, así como la generación de respuestas falsas

o sin sentido ("alucinación"). Estas limitaciones no son comportamientos anómalos, sino consecuencias inherentes al funcionamiento de estos sistemas, que se pueden mitigar hasta cierto punto con diversas técnicas (de las cuales hemos citado el ajuste de instrucciones) pero, al menos de momento, sin garantías de eliminarlas por completo.

Por otra parte, y siendo conscientes de estas limitaciones cuando los usamos, los *LLMs* resultan muy útiles dado su manejo del idioma y su enorme versatilidad, siendo capaces de realizar una gran cantidad de tareas que involucran textos. Esto incluye la generación de respuestas que, para un ser humano, requerirían habilidades como creatividad, razonamiento lógico o conocimiento de la realidad física. Aunque es debatible hasta qué punto se puede decir que los *LLMs* poseen realmente estas capacidades o si solo actúan como loros imitadores y cualquier muestra aparente de creatividad o razonamiento en sus respuestas es una ilusión; lo cierto es que desde un punto de vista práctico, el resultado final a menudo es indistinguible de si las tuviesen.

No obstante, no debemos perder de vista que ahora mismo no sabemos cómo surgen las capacidades emergentes de los *LLMs,* y esto hace que su funcionamiento nos resulte muy opaco. Por lo tanto, y teniendo en cuenta también las observaciones previas sobre sesgos y alucinaciones, nunca se debe confiar en un *LLM* para tomar una decisión relevante.

Por último, cabe puntualizar que este capítulo se ha centrado en tratar de arrojar luz sobre los aspectos del debate público que hacen referencia a sus capacidades y limitaciones, que se derivan directamente de su funcionamiento. Quedan fuera del alcance de este capítulo cuestiones que escapan a los aspectos técnicos de los *LLMs*, y se centran más bien en consideraciones sociales, éticas, económicas o legales de su utilización: por ejemplo, las implicaciones legales de usar material protegido por copyright para entrenar *LLMs* sin permiso de los autores; los riesgos creados por su uso para generar información falsa, manipular a personas o cometer fraude académico; la posibilidad de que los *LLMs* acaben sustituyendo masivamente puestos de trabajo; la huella de carbono que generan debido a sus considerables requisitos computacionales; o el oligopolio que las grandes empresas tecnológicas ostentan sobre los mejores modelos. Si bien la discusión detallada de estos debates queda fuera de este capítulo, confío en que el análisis aquí presentado pueda contribuir también a abordar estas cuestiones en otros contextos.